\def\year{2021}\relax
\documentclass[letterpaper]{article} 
\usepackage{aaai21}  

\usepackage{multirow}

\usepackage{amsmath}

\usepackage{times}  
\usepackage{helvet} 
\usepackage{courier}  
\usepackage[hyphens]{url}  
\usepackage{graphicx} 
\usepackage{natbib}  
\usepackage{caption} 
\title{Regularization via Adaptive Pairwise Label Smoothing}
\author{
  Hongyu Guo \\
  } 
  \affiliations{
  National Research Council of Canada\\
  1200 Montreal Road, Ottawa \\
  \texttt{hongyu.guo@nrc-cnrc.gc.ca} \\
}

\begin{document}

\maketitle

\begin{abstract}

Label Smoothing (LS) is  an effective regularizer to improve the generalization  of   state-of-the-art deep models.  For each training sample  the LS strategy smooths the one-hot encoded training signal by distributing its distribution mass over the non ground-truth classes,  aiming to penalize the networks from generating overconfident output distributions.  This paper introduces a novel label smoothing technique called Pairwise Label Smoothing (PLS). The PLS  takes a pair of samples as input. Smoothing with a pair of ground-truth labels enables the  PLS to  preserve the relative distance between the two truth labels  while further soften that between the  truth labels  and the other targets, resulting in models producing much less confident predictions than the LS strategy. Also, unlike current LS methods, which typically require to  find a global smoothing distribution mass through cross-validation search, PLS  automatically  learns the   distribution mass for each input pair during training. We empirically show  that  PLS  significantly outperforms LS and the baseline models, achieving up to 30\% of relative  classification error reduction.  We also visually show that when achieving such accuracy gains the PLS  tends to  produce very low winning softmax scores.
\end{abstract}

\section{Introduction}
Label Smoothing (LS) is a commonly used output distribution regularization technique to improve the generalization performance of deep learning models~\citep{
Szegedy2016RethinkingTI,Chorowski2016TowardsBD,Vaswani2017AttentionIA,8579005,Real2019RegularizedEF,Huang2019GPipeET,structurels}. 
Instead of training with data associated with one-hot labels, models with label 
smoothing  are trained on samples with  soft targets, where each target is  
a weighted mixture of the ground-truth one-hot label  with the uniform distribution of the classes. Such  regularization approaches prevent  overfitting of a model by penalizing overconfident output distributions, resulting in improved model generalization and model calibration~\citep{Szegedy2016RethinkingTI,DBLP:conf/iclr/PereyraTCKH17,yuan2019revisit,MullerKH19,DBLP:conf/aaai/ZhuJZGHSZ20,Lukasik2020DoesLS}. Owing to its simplicity and effectiveness, label  smoothing  has been successfully deployed to improve the accuracy of deep models across a range of applications, including  image classification~\citep{Szegedy2016RethinkingTI}, speech recognition~\citep{Vaswani2017AttentionIA}, and machine translation~\citep{Chorowski2016TowardsBD}.

When distributing the distribution mass of the one-hot  training signal over the non ground-truth classes,   current smoothing strategies, however, only consider the distance between the only gold label and the other non-ground-truth targets. Our observations suggest that providing an additional  ground-truth label to constrain the smoothing strength can significantly improve the smoothing effectiveness. The additional ground-truth label  here acts as a  consistency constraint between the pair of ground-truth labels while distancing them from the other targets. Motivated by this observation, in this paper, we introduce a novel label smoothing technique that takes a pair of samples as input.  We term it Pairwise Label Smoothing (denoted as PLS). 

\begin{figure} [h]
\caption{Illustration of label smoothing (left) and pairwise label smoothing (right) on how the distribution mass of the one-hot ground-truth label(s) (in blue bars) are distributed over the non ground-truth targets (in red pattern bars) for a training sample; orange  solid bars are the resulting smoothed ground-truth targets. 
 Y-axis indicates the distribution mass of the  four target classes (X-axis) of the training sample. 
 }
\label{fig:ill}
\centering
\includegraphics[width=3.316in]{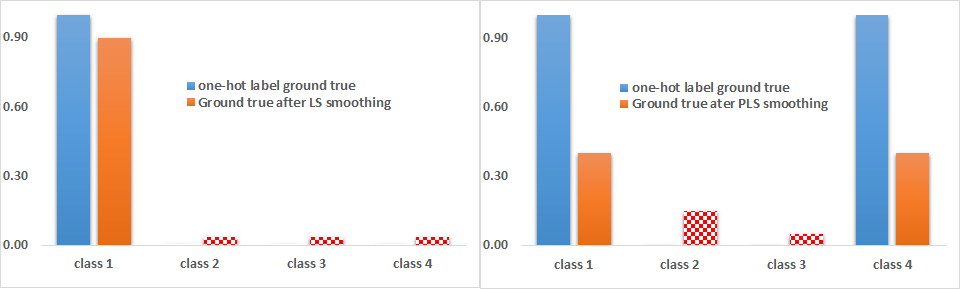}

\end{figure}

In a nutshell, the PLS first averages the inputs and labels of a pair of samples to form a new sample,  and then distributes the  distribution mass of the two ground-truth  targets  of the new sample over its non ground-truth classes. As illustrated in Figure~\ref{fig:ill}, smoothing with a pair  of ground-truth labels  enables  PLS to  preserve  the relative distance between the two truth labels  while being able to further soften that between the truth labels  and the other class targets. Also,   PLS  automatically  learns the  smoothing distribution mass for each input pair during training. Consequently, it effectively eliminates the   turning efforts for searching the right level of smoothing strength when applying  to different data and network architectures.

We empirically show that 
the PLS strategy significantly 
outperforms  LS  and the baseline models, 
with up to 30\% of relative classification error reduction. 
In addition,  we  visually demonstrate that PLS produces extremely conservative predictions  in testing time, resulting in that many of its  winning predicted softmax scores are slightly over 0.5. These low softmax scores come with  large Expected Calibration Error (ECE)~\citep{10.5555/3305381.3305518}, but such miscalibration error can be effectively reduced  by 
the post-training calibration  step Temperature Scaling~\citep{10.5555/3305381.3305518}. 
 
 Our contributions are as follows.  
\begin{itemize}  
   \item
We introduce a novel label smoothing method PLS, which leverages a pair of  ground-truth labels to constrain the smoothing strength and embraces  dynamic smoothing distribution. 
    \item
We experimentally demonstrate that the  PLS method   significantly improves the predictive accuracy  
    of deep classification networks. 
We also visually show that when achieving such  accuracy gains the PLS strategy tends to produce very low winning softmax scores. 
\end{itemize}

\section{Pairwise Label Smoothing}
\subsection{Preliminaries}
We consider a standard classification setting with a given training data set $(X; Y)$ associated with $K$ candidate classes $\{1,2,\cdots,K\}$. 
For an example $x_i$ from the training dataset $(X; Y)$, we denote the ground truth distribution $q$ over the labels as $q(y|x_i)$ ( $\sum_{y=1}^{K} q(y|x_i) = 1$). Also, we denote  a neural network  model to be trained as $f_{\theta}$ (parameterized with $\theta$), and it produces 
a conditional  label distribution over the $K$ classes as $p_{\theta}(y|x_i)$:
\begin{equation}
p_{\theta}(y|x_i) = \frac{\exp(z_{y})}{\sum_{k=1}^{K}\exp(z_{y_{k}})}, 
\end{equation}
with 
 $\sum_{y=1}^{K} p_{\theta}(y|x_i) = 1$, and $z$
is noted as the logit of the model $f_{\theta}$. The logits are generated with two steps: 
 the model $f_{\theta}$ first  constructs the $m$-dimensional input embedding $S_{i} \in R^{m}$ for the given input $x_{i}$, and then passes it 
  through
a linear   fullyconnected layer $W_{l} \in R^{K \times m}$: 
\begin{equation}
\label{logit}
S_{i} = f_{\theta}(x_{i}),
\end{equation}
\begin{equation}
\label{mmix}
z=  W_{l}S_{i}.
\end{equation}

During learning, the model $f_{\theta}$ is trained to optimize the parameter $\theta$ using the $n$ examples from  $(X; Y)$ by  minimizing the cross-entropy loss: 
\begin{equation}
\ell =- \sum_{i=1}^{n} H_i(q,p_{\theta}). 
\end{equation} 
Instead of using one-hot encoded vector for each example $x_{i}$ in $(X; Y)$, label smoothing (LS) adds a smoothed label distribution  (i.e., the prior distribution) $u(y|x_{i})$ to each example $x_{i}$, forming a new target label, namely  soft label:  
\begin{equation}
\label{oriequ}
q^{\prime}(y|x_i) = (1-\alpha) q(y|x_i) + \alpha u(y|x_i), 
\end{equation}
where hyper-parameter $\alpha$ is a weight factor ($\alpha \in [0,1]$) needed to be tuned to indicate the smoothing strength for the one-hot label. This modification results in a new loss function:
\begin{equation}
\label{lsloss}
\ell^{\prime} 
= - \sum_{i=1}^{n} \Big[ (1-\alpha) H_i(q,p_{\theta}) + \alpha H_i(u,p_{\theta}) \Big]. 
\end{equation} 
Usually, the $u(y|x_{i})$ is an uniform distribution, independent of data $x_{i}$, as $u(y|x_{i}) = 1/K$, and hyper-parameter $\alpha$ is tuned with cross-validation.

\subsection{Pairwise Label Smoothing}
\label{plsmethod}
Our proposed pairwise label smoothing method PLS leverages a pair of samples,  randomly selected from  $(X; Y)$, to conduct label smoothing. 

In detail, for a sample $x_{i}$ from the provided training  set $(X; Y)$ for training, PLS first randomly selects~\footnote{For efficiency purpose, we implement this by randomly selecting a sample from the same mini-batch during training.} another training sample $x_{j}$. For the  pair of samples $(x_{i}; y_{i})$ and $(x_{j} ; y_{j})$, where $x$ is the
input and $y$ the one-hot encoding of the corresponding class, PLS then generates a synthetic sample  through element-wisely averaging both the input features and the labels, respectively, as follows:  
\begin{equation}
x_{ij} =  (x_{i} +  x_{j})/2, 
\end{equation}
\begin{equation}
\label{equ:2}
q(y|x_{ij}) = ( y_{i} +  y_{j})/2.  
\end{equation}
In this way, for the sample $x_{ij}$ we have the ground truth distribution $q$ over the labels as $q(y|x_{ij})$ ($\sum_{y=1}^{K} q(y|x_{ij}) = 1$). 
The newly resulting  sample $x_{ij}$ will  then be  used for label smoothing (will be discussed in detail later) before feeding into the networks for training. 
In other words, the logits as defined in Equation~\ref{mmix} is computed by first generating the  $m$-dimensional image embedding $S_{ij} \in R^{m}$ for the input $x_{ij}$ and then passing through the fully-connected linear layer to construct the logit $z$: 
\begin{equation}
S_{ij} = f_{\theta}(x_{ij}), 
\end{equation}
\begin{equation}
\label{mmix22}
z= W_{l}S_{ij}.
\end{equation}
Similarly, the conditional  label distribution over the $K$ classes is computed as:
\begin{equation}
p_{\theta}(y|x_{ij}) = \frac{\exp(z_{y})}{\sum_{k=1}^{K}\exp(z_{y_{k}})}. \end{equation}

\subsubsection 
{Dynamic Smoothing Distribution}\label{evovinglabel} 
PLS leverages a learned distribution,  which depends on the input $x$, to dynamically generate the smoothing distribution mass for distributing the ground-truth target distribution to the non-target classes. 
To this end, the PLS implements this by adding a  fullyconnected layer  to the network $f_{\theta}$. That is,   the 
$f_{\theta}$ produces two projections from the penultimate layer representations of the network:  one for the logits as the original network (Equation~\ref{mmix22}), and another for generating the smoothing distribution as follows. 

In specific, an additional  fullyconnected
layer $W_{t} \in R^{K \times m}$ is added to the original networks $f_{\theta}$ to produce the smoothing distribution over the $K$ classification classes. 
That is, for the given 
input image $x_{ij}$,  its smoothing  distributions over the $K$ classification  targets, denoted as  $u^{\prime}_{\theta}(y|x_{ij})$, are computed as follows:  
\begin{equation}
\label{prelabel}
u^{\prime}_{\theta}(y|x_{ij}) = 
\frac{\exp(v_{y})}{\sum_{k=1}^{K}\exp(v_{y_{k}})},  
\end{equation}
   \begin{equation}
   \label{sha}
   v = \sigma (W_{t}S_{ij}), 
   \end{equation}
where $\sigma$ denotes the Sigmoid function, and  
 $S_{ij}$ is the same image embedding as that in Equation~\ref{mmix22}. In other words, the two predictions (i.e., Equations~\ref{mmix22} and~\ref{sha}) share the same networks except the last fully connected layer. 
That is, the only difference between PLS and the original networks is the added  fullyconnected layer $W_{t}$. 

After having the smoothing distributions $u^{\prime}_{\theta}(y|x_{ij})$, PLS then uses them to smooth the ground-truth labels  $q(y|x_{ij})$ as described in Equation~\ref{equ:2}, with an  average:  
\begin{equation}
\label{merge}
q^{\prime}(y|x_{ij}) = ( q(y|x_{ij}) +  u^{\prime}_{\theta}(y|x_{ij}))/2. 
\end{equation}

The loss function of PLS thus becomes the follows:
\begin{equation}
\begin{split}
\ell^{\prime} 
=& - \sum_{i=1}^{n} \Big[ 0.5 \cdot  H_i(q(y|x_{ij}),p_{\theta}(y|x_{ij})) \\
&+  0.5\cdot H_i(u^{\prime}_{\theta}(y|x_{ij}),p_{\theta}(y|x_{ij})) 
\Big].
\end{split}
\end{equation} 
For training, PLS minimizes, with gradient descent on mini-batch, the  loss  $\ell^{\prime}$. 
One more issue needed to be addressed for the training. That is, the data samples used for training, namely $(x_{ij}; y)$ may lack  information on the original training samples  $(x_{i}; y)$ due to the average operation in the PLS. To compensate this fact, 
we alternatively feed inputs to the networks with either a mini-batch from the original images, i.e., ${x}_{i}$ or ${x}_{ii}$,  or a mini-batch from the averaged images, i.e., ${x}_{ij}$. 
Note that, when training with the former, the networks still  need to learn to assign the smoothing distribution $ u^{\prime}_{\theta}(y|x_{ii})$ to form the  soft  targets $q^{\prime}(y|x_{ii})$ for the sample ${x}_{ii}$. 
As will be shown in the experiment  section, this training strategy is   important to PLS'  regularization effect.

\section{Experiments}
\subsection{Datasets, Baselines, and Settings}

We evaluate our proposed  method PLS with the following five benchmark image classification tasks.  
\noindent \textbf{MNIST} is a digit (1-10)  recognition dataset with 60,000 training and 10,000 test 28x28-dimensional gray-level images. 
\noindent 
\textbf {Fashion} is an image recognition dataset with the same scale as
MNIST, containing 10 classes of fashion product pictures.
\noindent 
\textbf{SVHN} is the Google street view house numbers recognition data set. It has 73,257 digits, 32x32 color images for training, 26,032 for testing, and 531,131 additional, easier samples. Following literature, we did not use the additional images.
\noindent 
\textbf{Cifar10} is an image classification task with  10 classes. It has 
  50,000 training  and 10,000 test samples. 
\noindent 
\textbf{Cifar100} is similar to Cifar10 but with 100 classes and 600 images each.

We conduct experiments  using the popular benchmarking networks  PreAct ResNet-18~\citep{DBLP:journals/corr/HeZR016} and ResNet-50~\citep{DBLP:journals/corr/HeZR016}. We also perform ablation evaluations using deeper and wider networks   
 WideResNet-28-10~\citep{DBLP:conf/bmvc/ZagoruykoK16} and 
  DenseNet-121~\citep{densenet}. 
We  compare with the state-of-the-art label smoothing methods~\citep{MullerKH19,Lukasik2020DoesLS} (denoted as ULS) with various smoothing coefficients (i.e., $\alpha$ as defined in Equation~\ref{lsloss}), where ULS-0.1, ULS-0.2, and ULS-0.3  denote the smoothing coefficient of 0.1, 0.2, and 0.3, respectively. 
We also compare our method with the input-pair based data augmentation method  Mixup~\citep{Mixup17}, which learns from synthetic samples generated from a pair of inputs through interpolations on both the inputs and labels of the input pair. For Mixup, we use the authors' code at~\footnote{https://github.com/facebookresearch/mixup-cifar10} and the uniformly selected mixing coefficients between [0,1]. 
For PreAct ResNet-18, ResNet-50, and DenseNet-121, we use the PyTorch implementation  from Facebook~\footnote{https://github.com/facebookresearch/mixup-cifar10/blob/master/models/}. For PLS, the added fullyconnected layer is the same as  the last fullyconnected layer of the baseline network with a Sigmoid function on the top. 
All models  are trained using mini-batched (128 examples) backprop, 
 with the  exact settings as in the Facebook codes,     for 400 epochs.   Each reported  value (accuracy or error rate) is the mean of   five runs. All our experiments were run on a NVIDIA GTX TitanX GPU with 12GB memory. 

\begin{table*}[h]
  \centering
\scalebox{0.9913}{
\begin{tabular}{l|c|c|c|c|c}\hline
Methods& MNIST&Fashion&SVHN&Cifar10&Cifar100\\ \hline
PreAct ResNet-18 &0.62	$\pm$0.05&4.78$\pm$	0.19&3.64$\pm$	0.42&5.19$\pm$	0.30&	24.19$\pm$	1.27\\
ULS-0.1 &0.63$\pm$0.02&4.81$\pm$0.07&3.20 $\pm$ 0.06&4.95$\pm$0.15&21.62$\pm$0.29\\
ULS-0.2 &0.62$\pm$0.02&4.57$\pm$0.05&3.14 $\pm$ 0.11&4.89$\pm$0.11&21.51$\pm$0.25\\
ULS-0.3 &0.60$\pm$0.01&4.60$\pm$0.06&3.12 $\pm$ 0.03&5.02$\pm$0.12&21.64$\pm$0.27\\
Mixup  &0.56$\pm$	0.01&4.18$\pm$	0.02&3.37$\pm$	0.49&3.88$\pm$	0.32&	21.10$\pm$	0.21\\
\hline
PLS &\textbf{0.47 $\pm$0.03}&\textbf{3.96$\pm$0.05}&\textbf{2.68 $\pm$0.09}&\textbf{3.63$\pm$0.10}&\textbf{19.14$\pm$0.20}\\
Rel. Imp. over Ave. ULS (\%)&23.78&15.02&15.01&26.72&11.35\\
Rel. Imp. over Baseline  (\%)&24.19&17.15&26.37&30.06&20.88\\
\hline
\end{tabular}
}
\caption{Error rate (\%) of the testing methods with PreAct ResNet-18~\citep{DBLP:journals/corr/HeZR016} as baseline.  We report mean scores over 5 runs with standard deviations (denoted $\pm$). The relative improvement of PLS over the baseline  and the average of the label smoothing ULS with various smoothing coefficients are  provided in the last two rows of the table. Best results are  in \textbf{Bold}. 
  }    \label{tab:accuracy:resnet18} 
\end{table*} 

\begin{table*}[h]
  \centering
\scalebox{0.9937}{
\begin{tabular}{l|c|c|c|c|c}\hline
Methods& MNIST&Fashion&SVHN&Cifar10&Cifar100\\ \hline
ResNet-50 &0.61$\pm$0.05
	&4.55$\pm$0.14	&3.22$\pm$0.05&4.83$\pm$0.30&23.10$\pm$0.62	\\
ULS-0.1 &0.63$\pm$0.02&4.58$\pm$0.16&2.98$\pm$0.02&4.98$\pm$0.25&23.90$\pm$0.99\\
ULS-0.2 &0.62$\pm$0.03&4.52$\pm$0.04&3.08$\pm$0.03&5.00$\pm$0.35&23.88$\pm$0.73\\
ULS-0.3 &0.65$\pm$0.03&4.51$\pm$0.15&3.04$\pm$0.07&5.16$\pm$0.16&23.17$\pm$0.50\\
Mixup  &0.57$\pm$	0.03&4.31$\pm$0.05	&2.85$\pm$0.07&4.29$\pm$0.28&19.48$\pm$0.48\\
\hline
PLS &\textbf{0.51$\pm$
0.02}&\textbf{4.15$\pm$0.09}&\textbf{2.36$\pm$0.03}&\textbf{3.60$\pm$0.18}&\textbf{18.65$\pm$1.08}\\
Rel. Imp. over ULS (\%)&19.47&	8.52&	22.20&	28.67&	21.14\\
Rel. Imp. over Baseline  (\%)&16.39&8.79&	26.71&	25.47&	19.26\\

\hline
\end{tabular}
}
\caption{Error rate (\%) of the testing methods with  ResNet-50~\citep{DBLP:journals/corr/HeZR016} as baseline.  We report mean scores over 5 runs with standard deviations (denoted $\pm$). The relative improvement of PLS over the baseline model and the average of the label smoothing (ULS) with various smoothing coefficients are  provided in the last two rows of the table. Best results are  in \textbf{Bold}. 
  }    \label{tab:accuracy:resnet50} 
\end{table*} 

\subsection{Predictive Accuracy} 
The predictive error rates obtained by label smoothing with various smoothing coefficients (i.e., ULS), Mixup, and PLS with   PreAct ResNet-18  as baseline on the five test datasets are presented in Table~\ref{tab:accuracy:resnet18}, where the relative improvement of PLS over the baseline model PreAct ResNet-18 and the average of the three  ULS models (i.e., ULS-0.1, ULS-0.2, ULS-0.3) are in the last two rows of the table. The  results with ResNet-50 as baselines for the testing methods are provided in Table~\ref{tab:accuracy:resnet50}. 

The results in Table~\ref{tab:accuracy:resnet18} show that PLS outperforms, in terms of predictive error, the PreAct ResNet-18 baseline, the label smoothing models (ULS-0.1, ULS-0.2, ULS-0.3), and Mixup  on  all the five datasets. 
For example, the relative improvement of PLS over the baseline model  on the Cifar10 and MNIST datasets, as depicted in the  last row of the table, are over 30\% and 24\%, respectively. 
When considering PLS and the average error obtained by the three ULS models, as depicted in the second last row of the table, the relative error reduction on all the five datasets is at least 11\%; on both the Cifar10 and MNIST tasks, the relative improvement is over 23\%.


When considering the cases with ResNet-50 as baselines, results as shown in 
Table~\ref{tab:accuracy:resnet50}  indicate that similar error reductions are obtained by PLS. Again, on all the five testing datasets, PLS outperforms all the comparison baselines, namely ResNet-50, label smoothing with three different coefficient settings, and Mixup. Also, the relative improvement is large in some cases. For example, for the Cifar10 dataset, the relative improvement achieved by PLS over the baseline ResNet-50 and the average of the three label smoothing strategies (i.e. ULS) are 25.47\% and 28.67\%, respectively. 

These results suggest that the PLS method is able to significantly reduce the predictive error of the baseline models with or without label smoothing (ULS) applied. 


\subsubsection{Accuracy on Deeper and Wider Networks} 
We also evaluate PLS using deeper and wider networks:  WideResNet-28-10~\citep{DBLP:conf/bmvc/ZagoruykoK16}, and  DenseNet-121~\citep{densenet}. 
The results are in Table~\ref{tab:accuracy:wideresnet}. 

Similar to that of using PreAct ResNet-18 and ResNet-50 as presented in Tables~\ref{tab:accuracy:resnet18} and~\ref{tab:accuracy:resnet50}, PLS  performed consistently better on the two additional network architectures, i.e.,  the much wider network WideResNet-28-10  and the 121 layers network DenseNet-121. For example, as shown in the table, PLS significantly outperformed the  baseline and the label smoothing strategy ULS with different smoothing coefficients.

\begin{table}[h]
  \centering
\scalebox{0.97}{
 \begin{tabular}{l|c|c}\hline
Networks&WideResNet-28-10&DenseNet-121\\ \hline
Baseline &19.90	$\pm$ 0.41&19.50$\pm$	0.34\\
ULS-0.1 &19.59$\pm$0.17&19.42$\pm$0.10\\
ULS-0.2 &20.05$\pm$0.43&19.37$\pm$0.15\\
ULS-0.3 &20.41$\pm$0.28&19.45$\pm$0.09\\
Mixup &	18.11$\pm$	0.30&17.92$\pm$0.12	\\
\hline 
PLS &\textbf{17.12$\pm$0.26}&\textbf{16.90$\pm$0.22}\\ 
\hline
\end{tabular}
}
\caption{Error rates (\%) of testing methods with WideResNet28-10 and DenseNet-121 as baselines. We report mean scores over 5 runs with standard deviations (denoted $\pm$).  Best results are in \textbf{Bold}.
  }  
  \label{tab:accuracy:wideresnet} 
\end{table}


\subsubsection{Training Characteristics}
\begin{figure}[h]
\caption{Training loss (right) and validation error (left) of PreAct ResNet-18 on  Cifar100. 
 }
\label{fig:converge}
\centering
\includegraphics[width=3.3in]{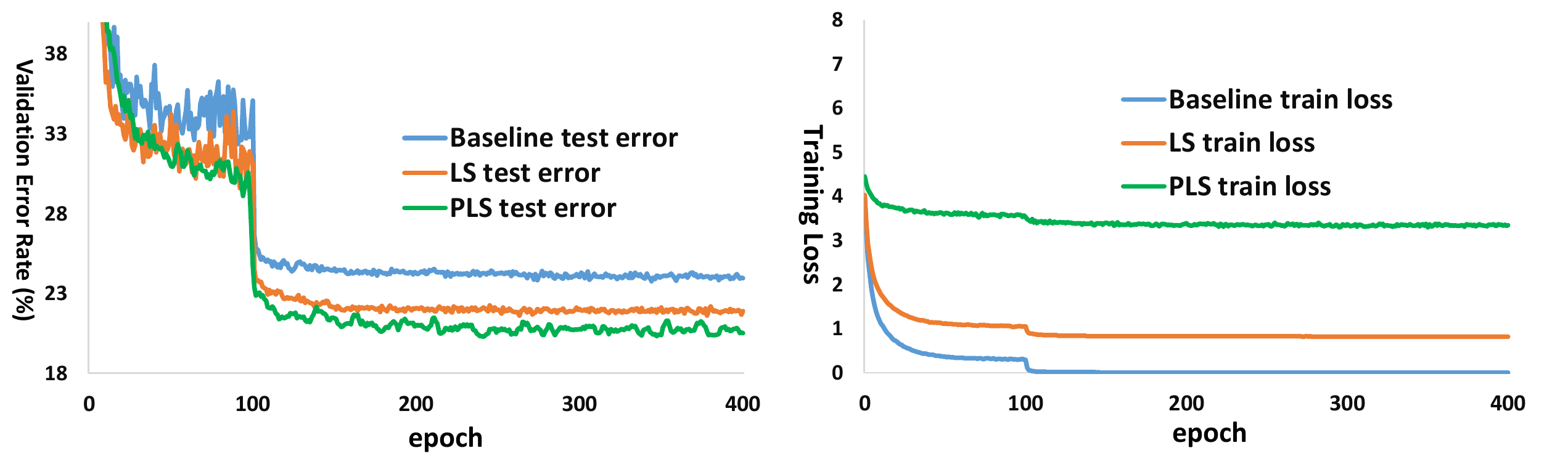}

\end{figure}

To better understand the training characteristics of the PLS method, in Figure~\ref{fig:converge} we  plot the training loss and validation error rate across the 400 training epochs of  PLS, ULS-0.1, and  PreAct ResNet-18  on the  Cifar100 dataset.   

 Figure~\ref{fig:converge} shows that  the baseline mode's training loss goes to zero after 100 epochs, providing no gradient after that.  On the contrary, the  training loss  of PLS (green curve in the right subfigure) maintains a relatively higher  level than the baseline and the ULS methods,  allowing the PLS model to keep tuning the networks. Promisingly, as shown by the  validation error curve (green curve in the left subfigure),  PLS  is not overfitting the training set with the high training loss but keep decreasing the  validation error.

\subsubsection{Impact of Key Components}
\label{ablation}
We  also evaluate the impact of the key components in  PLS using PreAct ResNet-18 and ResNet-50 on Cifar100. Results are in Table~\ref{aaaablation}. The key components include  1) removing the  learned smoothing distribution mass in Equation~\ref{merge},  2) excluding the use of the original training inputs as discussed in the method section, and 3) replacing the learned smoothing distribution mass with uniform distribution with weighting coefficients of 0.1, 0.2, and 0.3 (denoted as UD-0.1, UD-0.2 and UD-0.3).

The error rates obtained in  Table~\ref{aaaablation} show that, 
  both  the learned smoothing distribution mass and  the original training samples are critical for the PLS model. 
In particular, when excluding the original  samples  from  training, the predictive error of PLS  dramatically increased from about 19\% to nearly 24\% for both the PreAct ResNet-18 and ResNet-50. The reason,  as discussed in the method section, 
 is that,  
without the original training samples, the networks may lack  information on the validation samples.

\begin{table}[h]
  \centering
  \begin{tabular}{l|r|r}\hline
 \multirow{2}{*}{PLS }& ResNet-18&ResNet-50 \\ 
  &19.14&18.65 \\ \hline
----  no learned distribution  &21.06&19.35\\
----  no original  images &23.84&24.42\\ 
 \hline
----  UD 0.1&19.50&18.91\\
----  UD 0.2&19.25&18.81\\
----  UD 0.3&19.31&18.89
\\ 
\hline
\end{tabular}
  \caption{Error rates (\%) on Cifar100 obtained by PLS  while varying its key components.  
\label{aaaablation}}

\end{table}

Also, results  in Table~\ref{aaaablation} indicate that,  replacing the learned smoothing distribution mass in PLS with manually tuned Uniform distribution mass (i.e., UD) obtained slightly larger errors. Theses results indicate that the PLS method is able to adaptively learn the  distribution mass for smoothing the two target labels, resulting in superior  accuracy and excluding the need for the  coefficient search  for different  applications. 
These results here also further confirm the benefits of conducting label smoothing on sample pairs.


\begin{figure}[h]
\centering
    \includegraphics[width=2.20972in, height=1.002in]{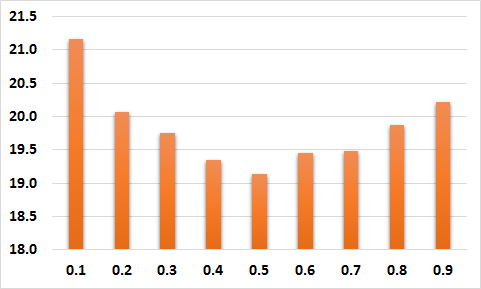}     
  \caption{
  Error rates on Cifar100 obtained by 
  varying the weight factor in PLS. } 
  \label{weighting}
\end{figure}

\subsubsection{Re-weight Smoothing Strength}
PLS distributes half of its ground-truth distribution mass over the non ground-truth targets as discussed in Equation~\ref{merge}. 
We here  evaluate the impact of different weight factors between the ground-truth and non ground-truth targets, by varying it from 0.1 to 0.9. Note that, 0.5 equals to the average used by PLS. The results obtained by PLS  using PreAct ResNet-18 on Cifar100 are in Figure~\ref{weighting}. The error rates obtained in Figure~\ref{weighting} suggest 
that average as deployed by PLS 
provides better accuracy than that of other weighting ratios. 
Why average works better is not obvious to us, so we would like to leave it to future work. 

\begin{figure}[h]
\caption{
Average distribution mass of ground-truth targets 
(green bar)
and the top 5 largest  non ground-truth targets (red bar) used by PLS in  training  with PreAct ResNet-18 on Cifar100 (left) and Cifar10 (right).  
X-axis depicts the targets, and Y-axis indicates their distribution mass. 
 }
\label{fig:top3}
\centering
\includegraphics[width=3.3in]{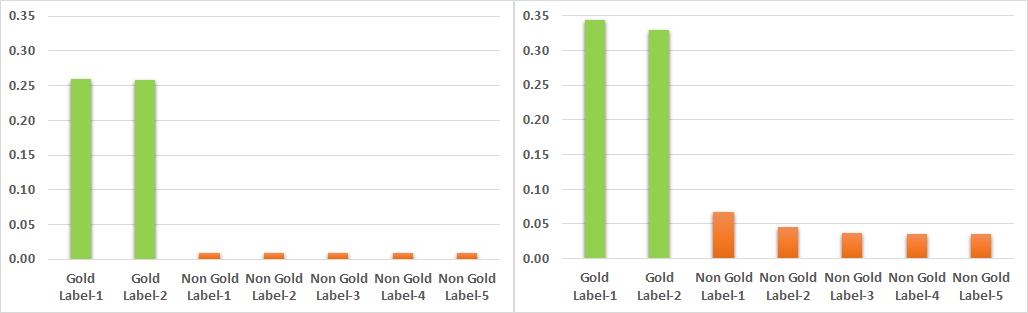}

\end{figure}

\vspace{-2mm}

\subsection{Why and How   PLS Benefits from Label Pair?}
\label{confidence}

\subsubsection{Soft Labels Used in Training} 
To peek into how the additional ground-truth label in PLS constrains the smoothing distribution mass,  
 we visualize, in Figure~\ref{fig:top3}, the soft target labels used for training by  PLS  with PreAct ResNet-18 on  Cifar100 (left) and Cifar10 (right). Figure~\ref{fig:top3} depict the   soft label values of the training samples for both the ground-truth targets (in green) and their top 5 largest  non ground-truth classes (in red). The figure presents the average values over  all the training samples resulting from sample pairs with two different one-hot true labels.  In the figure, the X-axis depicts the training targets, and the  Y-axis indicates the corresponding distribution mass. 
 
Results in Figure~\ref{fig:top3} indicate that, PLS uses much smaller target values for the ground-truth labels during training, when compared to the one-hot representation label used by the baseline models and the soft targets used by label smoothing ULS. 
For example, the largest ground-truth training targets for PLS are around  0.25 and 0.35  (green bars), respectively, for Cifar100 and Cifar10. These values are much smaller than the 1.0 used by the baseline models and the 0.9 used by the ULS-0.1 models. 
Consequently, in PLS, the distance between  the  ground-truth targets and the non ground-truth targets are much smaller than that in the baseline and the ULS-0.1 models. In addition, the distance between the two ground-truth targets in PLS (green bars) is very small, when compared to the distance between the ground-truth and non ground-truth target values (green vs. red bars). 

Theses results indicate that smoothing with a pair  of ground-truth labels  enables  PLS to  preserve  the relative distance between the two truth labels  while being able to further soften that between the truth labels  and the other class targets, for both  Cifar100 and Cifar10. 
In other words, the training samples in PLS have smoother training targets across all classes, and  those  training signals are far from 1.0, which in turn impacts how PLS makes its classification decisions as will be discussed next.  

\begin{figure}[h]
\caption{ Histograms of predicted softmax scores on the validation data   by PreAct ResNet-18 (top), ULS-0.1 (middle), and PLS (bottom) on Cifar100 (left) and Cifar10 (right). 
 }
	\centering
\includegraphics[width=3.4in,height=2.8in]{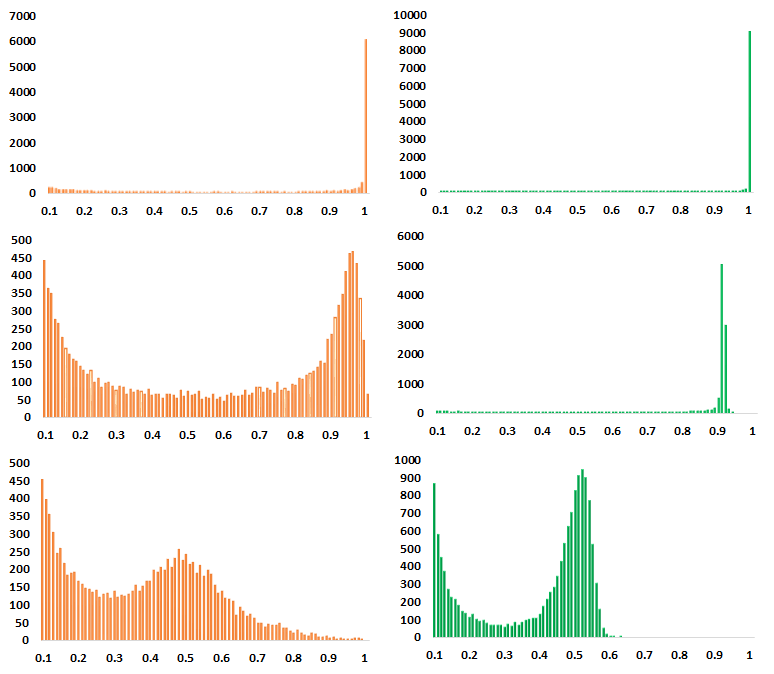}
	\label{fig:hist}
\end{figure}


\subsubsection{Predicted Softmax Scores in Testing} 
We observed that a direct effect of the low training signals (far from 1.0) as discussed above was reflected on the model's prediction scores made in test time. 
In this section, we present the PLS strategy's predicted softmax scores made in test time  in order to get further idea about why and how the PLS methods benefit from a pair of  samples when conducting label smoothing. 

In Figure~\ref{fig:hist}, we visualize the predicted softmax scores  made by PreAct ResNet-18 (top), ULS-0.1 (middle), and PLS (bottom) on all the 10K test data samples in Cifar100 (left column) and Cifar10 (right column). To have better visualization, we have removed the predictive scores less than 0.1 for all the testing methods since all models obtained similar results for confidences smaller than 0.1. 

For the Cifar100 dataset, results on the left of  Figure~\ref{fig:hist} indicate that  the baseline model PreAct ResNet-18 produced very confident predictions (top), namely skewing large mass of its predicted softmax scores on 1.0 (i.e., 100\% confidence). On the other hand,  the ULS method was able to 
decrease its prediction confidence at test time (middle). It  spreads its predictive confidences to the two ends, namely moving most of the predicted softmax scores into two bins [0.1-0.2] and [0.9-1.0]. 
Interestingly, the PLS strategy produced very conservative predicted softmax scores, by distributing many of its predicted scores to the middle, namely 0.5, and with sparse distribution for scores  larger than 0.7 (bottom subfigure).

When considering the Cifar10 data, results as shown on the right of Figure~\ref{fig:hist} again indicate that the baseline model PreAct ResNet-18 (top) produced very confident predictions, namely put large mass of its predicted softmax scores  near 1.0. 
For the ULS model (middle), the predicted softmax scores were also distributed near the 1.0, but it is much less than that of PreAct ResNet-18. On the other hand, the PLS method (bottom) again generated very conservative predicted softmax scores. The predicted scores mostly distribute near the middle point of the  softmax score range, namely 0.5, with a very few larger than 0.7. 

These results suggest that, resulting from the further smoothed  training target signals across classes, PLS becomes extremely conservative when generating predicted scores in test time, producing small winning softmax scores when making classification decisions.

\subsubsection{Model Calibration}
To further tease out the  conservative predictions preferred by the PLS method, we  evaluate how such conservative softmax scores  affect the calibration of the output probabilities. In this section, we report the Expected Calibration Error (ECE)~\citep{10.5555/3305381.3305518} obtained by the baseline PreAct ResNet-18, ULS-0.1, and PLS on the test set with 15 bins as used in~\citep{MullerKH19} for both  Cifar100 and Cifar10. 

\begin{figure}[h]
\caption{ ECE on  validation data of Cifar100 (left) and Cifar10 (right)  by  ResNet-18, ULS-0.1 and PLS. }
	\centering
\includegraphics[width=3.3in]{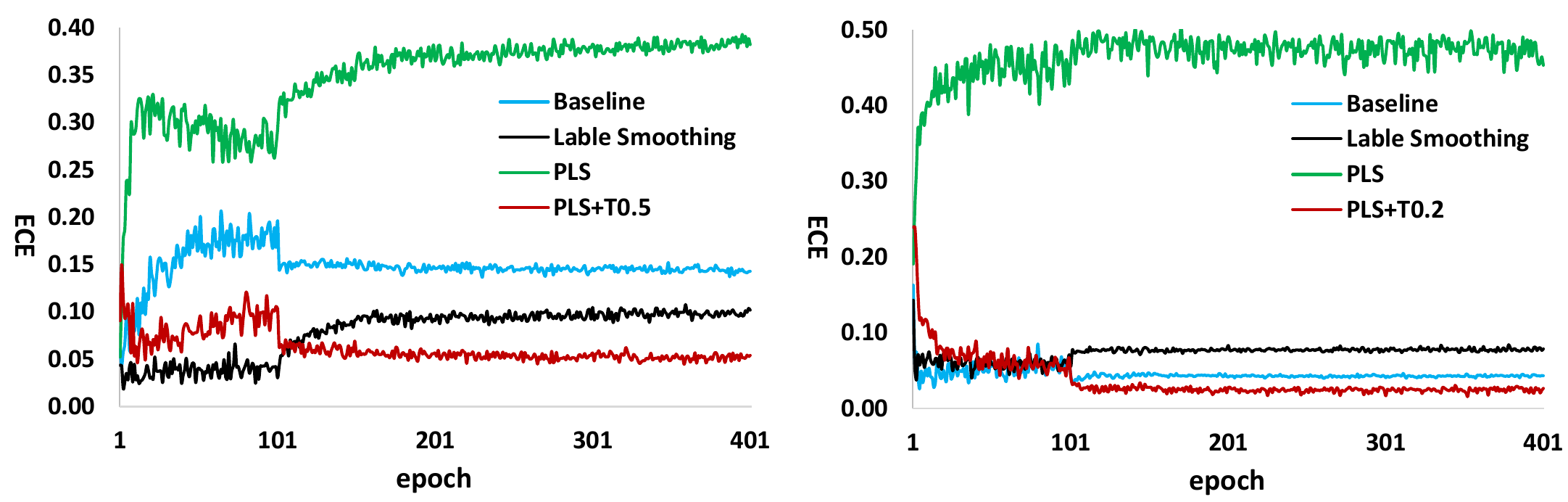}
	\label{fig:ece}
\end{figure}

Results in Figure~\ref{fig:ece} indicate that ULS (dark curve) is able to reduce the miscalibration error ECE on the Cifar100 data set (left subfigure), but for the Cifar10 dataset (right subfigure), ULS  has larger ECE error after 100 epochs of training than the baseline model. However, the ECE errors obtained by the PLS methods for both the Cifar100 and Cifar10 are much larger than  both the baseline and the ULS models. 
Note that, although the authors in~\citep{10.5555/3305381.3305518} state that the Batch Normalization (BN) strategy~\citep{DBLP:conf/nips/Ioffe17} also increases the miscalibration ECE errors for unknown reasons, we doubt that  the PLS will have the same reason as the BN approach. 
 This is because the main characteristic of the PLS model is that it produces extremely conservative winning softmax scores which is not the case for the BN strategy.  
We here suspect that the high ECE score of the PLS method may be caused by the fact that ECE is an evenly spaced binning metrics but the PLS produces sparse dispersion of the softmax scores  across the range. 
As previously shown in Figure~\ref{fig:hist}, PLS produces a very sparsely populated region 
 at the rightward end. 


To verify the above hypothesis, we further investigate the Temperature Scaling (TS) method introduced  in~\citep{10.5555/3305381.3305518}, which allows us to redistribute the distribution dispersion after training with no impact on the testing accuracy. 
During testing, TS  multiplies the logits by a scalar before applying the softmax operator. 
 We apply this TS technique to PLS, and present the results in Figure~\ref{fig:ece}, depicting by  red curve in the left and right subfigures for the Cifar100 and Cifar10, respectively. The TS factors was 0.5 and 0.2 respectively for Cifar100 and Cifar10, which were found by a search with 10 percentage of the training data. 
Results in Figure~\ref{fig:ece} indicate that the TS can significantly improve the  calibration of the PLS for both cases. The ECE errors obtained by PLS-TS for both the Cifar100 and Cifar10 cases (red curves) are lower than both the baseline (blue curves) and the label smoothing models ULS (black curves), and did not show any upward trend during training. 

\vspace{-2mm}
\begin{figure}[h]
\caption{ 
Numbers of samples (red bar) and ECE errors (green curve) in different bins used by ECE for the PLS (left) and PLS + Temperature Scaling (right) on Cifar100 (top row) and Cifar10 (bottom row). 
Note that the pre-set bin number is 15, but not all bins have samples. 
}
	\centering
\includegraphics[width=3.3in]{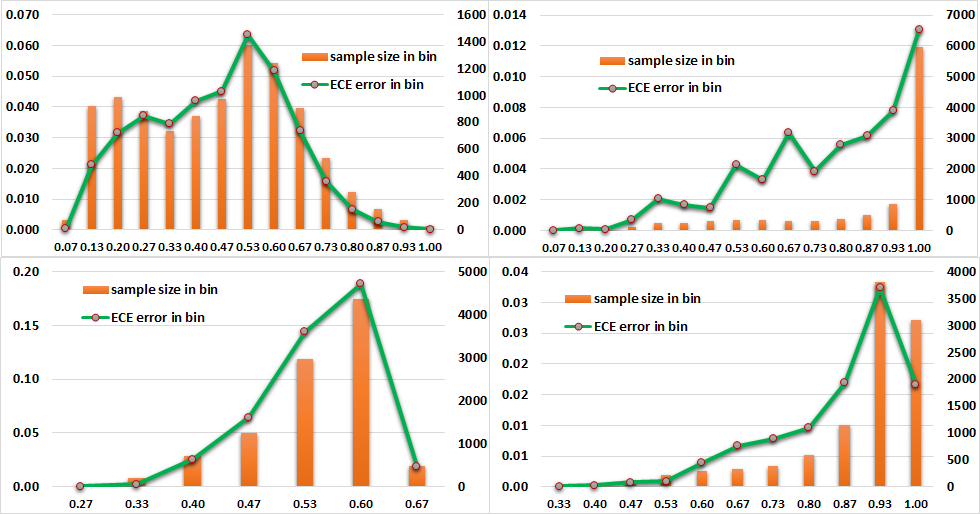}
	\label{fig:temperaturescaling}
\end{figure}

\vspace{-2mm}

How does TS help PLS in terms of ECE error? 
 Figure~\ref{fig:temperaturescaling} further depicts  the number of samples (right y-axis) for each of the 15 pre-set bins~\footnote{Note that not all bins have samples.} and its corresponding ECE error (left y-axis) without  and with  TS applied to PLS at the left and right subfigure respectively. We present the results for   Cifar100 and Cifar10 at the top and bottom rows, 
  respectively.  
As shown in Figure~\ref{fig:temperaturescaling}, 
 samples in the densely populated distribution regions in the middle (around 0.5), as depicted at the left subfigures (red bars), were  effectively redistributed by the TS to the sparsely populated area at the right end near 1.0, resulting in the redistributed mass in the right subfigures.  
As a result, the ECE error for each bin got smaller (green curve) after the redistribution by the TS. This effect resulted in smaller overall miscalibration error than the baseline and the ULS methods as presented previously. 
It is worth noting that, unlike typical networks, the TS factor is usually larger than 1.0 to avoid over-confident predictions~\citep{10.5555/3305381.3305518}, for PLS the scaling factor needs to be less than 1.0 in order to re-distribute
the densely populated distribution regions in the middle to the sparsely populated area at the right end.

\section{Related Work}
Label smoothing has shown to  provide consistent accuracy gains across many tasks~\citep{Szegedy2016RethinkingTI,DBLP:conf/iclr/PereyraTCKH17}. 
Müller et al.~\citep{MullerKH19}  explicitly show that label smoothing not only  improves model generalization but also enhances model calibration, empowering the models to generate less over-confident predictions. Li et al.~\citep{Lukasik2020DoesLS}  demonstrate that label smoothing can also  mitigate label noise. In~\citep{structurels}, Li et al.  
identify a quantifiable bias of the Bayes error rate in the uniform label smoothing, and propose a cluster-based strategy to fix that. 
Label smoothing also relates to DisturbLabel~\citep{7780883}, which can be seen as a marginalized version of label dropout.
Unlike the above methods which apply label smoothing to  each single input, our smoothing strategy leverages a pair of inputs for label smoothing. Also, the smoothing distribution mass of our approach is not uniform but  dynamic, which is automatically computed based on the pair of inputs. 

Our work is also related to methods that leverage a pair of samples for learning such as Mixup~\citep{Mixup17} and  variants~\citep{DBLP:conf/iclr/TokozumeUH18,adamixup,cutmix19,li2020feature,attributemix,supermix,DBLP:conf/aaai/Guo20,attentioncutmix,DBLP:journals/corr/abs-1905-08941,DBLP:journals/corr/abs-1906-06875}. These methods focus on augmenting the size of the training data. They  interpolate both the features and labels through random mixing coefficients sampling  between [0,1]. 
Our work here does not create a set of synthetic inputs from a pair of inputs. 
 Instead, our method focuses on distributing the distribution mass of the two one-hot  labels of an input pair to the other non ground-truth targets of the tasks.  

The generation of the dynamic smoothing distribution mass in PLS is also related to how the training targets generated by self-distillation methods~\cite{hinton2015distilling,Furlanello2018BornAN,DBLP:conf/aaai/YangXQY19,ahn2019variational,mobahi2020selfdistillation}. However, these self-distillation approaches treat the final predictions as target labels for a new round of training, and  the teacher and student architectures are identical~\cite{mobahi2020selfdistillation}.  In PLS, the target  prediction network and the smoothing distribution mass generation network have different architectures, and the training targets for the classification model are a mix of the outputs of the two networks.

\section{Conclusion and Future Work}
We contributed a novel label smoothing strategy PLS. The PLS leverages a pair of inputs to constrain the distributing of the distribution mass of the ground-truth labels over  other non ground-truth targets.  
 We empirically showed that  PLS  significantly outperforms, in terms of predictive accuracy, the baseline models with or without uniform label smoothing applied. 
 We  visually showed that when achieving such accuracy gains the PLS strategy tends to produce very low winning softmax scores. 
We  also empirically demonstrated that these conservative predictions come with large  miscalibration error ECE, but  such ECE error can be effectively reduced by the simple post training process step Temperature Scaling. 

Our studies here suggest some  interesting directions for future investigation. For example, what are the other benefits for very conservative predicted softmax scores?  
Another interesting research direction would be providing theoretical explanation  on why the PLS and Batch Normalization strategies  improve the predictive accuracy but decrease the model miscalibration score ECE.

\bibliography{labelsmoothing}

\begin{thebibliography}{34}
\providecommand{\natexlab}[1]{#1}
\providecommand{\url}[1]{\texttt{#1}}
\providecommand{\urlprefix}{URL }
\expandafter\ifx\csname urlstyle\endcsname\relax
  \providecommand{\doi}[1]{doi:\discretionary{}{}{}#1}\else
  \providecommand{\doi}{doi:\discretionary{}{}{}\begingroup
  \urlstyle{rm}\Url}\fi

\bibitem[{Ahn et~al.(2019)Ahn, Hu, Damianou, Lawrence, and
  Dai}]{ahn2019variational}
Ahn, S.; Hu, S.~X.; Damianou, A.; Lawrence, N.~D.; and Dai, Z. 2019.
\newblock Variational Information Distillation for Knowledge Transfer.
\newblock In \emph{arXiv}.

\bibitem[{Archambault et~al.(2019)Archambault, Mao, Guo, and
  Zhang}]{DBLP:journals/corr/abs-1906-06875}
Archambault, G.~P.; Mao, Y.; Guo, H.; and Zhang, R. 2019.
\newblock MixUp as Directional Adversarial Training.
\newblock volume abs/1906.06875.

\bibitem[{Chorowski and Jaitly(2016)}]{Chorowski2016TowardsBD}
Chorowski, J.; and Jaitly, N. 2016.
\newblock Towards Better Decoding and Language Model Integration in Sequence to
  Sequence Models.
\newblock In \emph{INTERSPEECH}.

\bibitem[{Dabouei et~al.(2020)Dabouei, Soleymani, Taherkhani, and
  Nasrabadi}]{supermix}
Dabouei, A.; Soleymani, S.; Taherkhani, F.; and Nasrabadi, N.~M. 2020.
\newblock SuperMix: Supervising the Mixing Data Augmentation.
\newblock \emph{arXiv preprint arXiv:2003.05034} .

\bibitem[{Furlanello et~al.(2018)Furlanello, Lipton, Tschannen, Itti, and
  Anandkumar}]{Furlanello2018BornAN}
Furlanello, T.; Lipton, Z.~C.; Tschannen, M.; Itti, L.; and Anandkumar, A.
  2018.
\newblock Born Again Neural Networks.
\newblock In \emph{ICML}.

\bibitem[{Guo et~al.(2017)Guo, Pleiss, Sun, and
  Weinberger}]{10.5555/3305381.3305518}
Guo, C.; Pleiss, G.; Sun, Y.; and Weinberger, K.~Q. 2017.
\newblock On Calibration of Modern Neural Networks.
\newblock In \emph{ICML}, ICML’17. JMLR.org.

\bibitem[{Guo(2020)}]{DBLP:conf/aaai/Guo20}
Guo, H. 2020.
\newblock Nonlinear Mixup: Out-Of-Manifold Data Augmentation for Text
  Classification.
\newblock In \emph{AAAI}, 4044--4051.

\bibitem[{Guo, Mao, and
  Zhang(2019{\natexlab{a}})}]{DBLP:journals/corr/abs-1905-08941}
Guo, H.; Mao, Y.; and Zhang, R. 2019{\natexlab{a}}.
\newblock Augmenting Data with Mixup for Sentence Classification: An Empirical
  Study.
\newblock volume abs/1905.08941.

\bibitem[{Guo, Mao, and Zhang(2019{\natexlab{b}})}]{adamixup}
Guo, H.; Mao, Y.; and Zhang, R. 2019{\natexlab{b}}.
\newblock MixUp as Locally Linear Out-of-Manifold Regularization.
\newblock In \emph{AAAI}, 3714--3722.

\bibitem[{He et~al.(2016)He, Zhang, Ren, and Sun}]{DBLP:journals/corr/HeZR016}
He, K.; Zhang, X.; Ren, S.; and Sun, J. 2016.
\newblock Identity Mappings in Deep Residual Networks.
\newblock \emph{ECCV} .

\bibitem[{Hinton, Vinyals, and Dean(2015)}]{hinton2015distilling}
Hinton, G.; Vinyals, O.; and Dean, J. 2015.
\newblock Distilling the Knowledge in a Neural Network.
\newblock In \emph{arXiv}.

\bibitem[{{Huang} et~al.(2017){Huang}, {Liu}, {Van Der Maaten}, and
  {Weinberger}}]{densenet}
{Huang}, G.; {Liu}, Z.; {Van Der Maaten}, L.; and {Weinberger}, K.~Q. 2017.
\newblock Densely Connected Convolutional Networks.
\newblock In \emph{2017 IEEE Conference on Computer Vision and Pattern
  Recognition (CVPR)}, 2261--2269.

\bibitem[{Huang et~al.(2019)Huang, Cheng, Chen, Lee, Ngiam, Le, and
  Chen}]{Huang2019GPipeET}
Huang, Y.; Cheng, Y.; Chen, D.; Lee, H.; Ngiam, J.; Le, Q.~V.; and Chen, Z.
  2019.
\newblock GPipe: Efficient Training of Giant Neural Networks using Pipeline
  Parallelism.
\newblock In \emph{NeurIPS}.

\bibitem[{Ioffe(2017)}]{DBLP:conf/nips/Ioffe17}
Ioffe, S. 2017.
\newblock Batch Renormalization: Towards Reducing Minibatch Dependence in
  Batch-Normalized Models.
\newblock In Guyon, I.; von Luxburg, U.; Bengio, S.; Wallach, H.~M.; Fergus,
  R.; Vishwanathan, S. V.~N.; and Garnett, R., eds., \emph{NeurIPS},
  1945--1953.

\bibitem[{Li et~al.(2020{\natexlab{a}})Li, Wu, Lim, Belongie, and
  Weinberger}]{li2020feature}
Li, B.; Wu, F.; Lim, S.-N.; Belongie, S.; and Weinberger, K.~Q.
  2020{\natexlab{a}}.
\newblock On Feature Normalization and Data Augmentation.
\newblock In \emph{arXiv}.

\bibitem[{Li et~al.(2020{\natexlab{b}})Li, Zhang, Xiong, and
  Tian}]{attributemix}
Li, H.; Zhang, X.; Xiong, H.; and Tian, Q. 2020{\natexlab{b}}.
\newblock Attribute Mix: Semantic Data Augmentation for Fine Grained
  Recognition.
\newblock In \emph{arXiv}.

\bibitem[{Li, Dasarathy, and Berisha(2020)}]{structurels}
Li, W.; Dasarathy, G.; and Berisha, V. 2020.
\newblock Regularization via Structural Label Smoothing.
\newblock In \emph{AISTAT}.

\bibitem[{Lukasik et~al.(2020)Lukasik, Bhojanapalli, Menon, and
  Kumar}]{Lukasik2020DoesLS}
Lukasik, M.; Bhojanapalli, S.; Menon, A.~K.; and Kumar, S. 2020.
\newblock Does label smoothing mitigate label noise?
\newblock \emph{ICML} .

\bibitem[{Mobahi, Farajtabar, and Bartlett(2020)}]{mobahi2020selfdistillation}
Mobahi, H.; Farajtabar, M.; and Bartlett, P.~L. 2020.
\newblock Self-Distillation Amplifies Regularization in Hilbert Space.
\newblock In \emph{arXiv}.

\bibitem[{M{\"{u}}ller, Kornblith, and Hinton(2019)}]{MullerKH19}
M{\"{u}}ller, R.; Kornblith, S.; and Hinton, G.~E. 2019.
\newblock When does label smoothing help?
\newblock In \emph{NIPS}.

\bibitem[{Pereyra et~al.(2017)Pereyra, Tucker, Chorowski, Kaiser, and
  Hinton}]{DBLP:conf/iclr/PereyraTCKH17}
Pereyra, G.; Tucker, G.; Chorowski, J.; Kaiser, L.; and Hinton, G.~E. 2017.
\newblock Regularizing Neural Networks by Penalizing Confident Output
  Distributions.
\newblock In \emph{ICLR workshop}.

\bibitem[{Real et~al.(2019)Real, Aggarwal, Huang, and
  Le}]{Real2019RegularizedEF}
Real, E.; Aggarwal, A.; Huang, Y.; and Le, Q.~V. 2019.
\newblock Regularized Evolution for Image Classifier Architecture Search.
\newblock In \emph{AAAI}.

\bibitem[{Szegedy et~al.(2016)Szegedy, Vanhoucke, Ioffe, Shlens, and
  Wojna}]{Szegedy2016RethinkingTI}
Szegedy, C.; Vanhoucke, V.; Ioffe, S.; Shlens, J.; and Wojna, Z. 2016.
\newblock Rethinking the Inception Architecture for Computer Vision.
\newblock \emph{2016 IEEE Conference on Computer Vision and Pattern Recognition
  (CVPR)} 2818--2826.

\bibitem[{Tokozume, Ushiku, and Harada(2018)}]{DBLP:conf/iclr/TokozumeUH18}
Tokozume, Y.; Ushiku, Y.; and Harada, T. 2018.
\newblock Learning from Between-class Examples for Deep Sound Recognition.
\newblock In \emph{ICLR}.

\bibitem[{Vaswani et~al.(2017)Vaswani, Shazeer, Parmar, Uszkoreit, Jones,
  Gomez, Kaiser, and Polosukhin}]{Vaswani2017AttentionIA}
Vaswani, A.; Shazeer, N.; Parmar, N.; Uszkoreit, J.; Jones, L.; Gomez, A.~N.;
  Kaiser, L.; and Polosukhin, I. 2017.
\newblock Attention is All you Need.
\newblock \emph{ArXiv} abs/1706.03762.

\bibitem[{Walawalkar et~al.(2020)Walawalkar, Shen, Liu, and
  Savvides}]{attentioncutmix}
Walawalkar, D.; Shen, Z.; Liu, Z.; and Savvides, M. 2020.
\newblock Attentive CutMix: An Enhanced Data Augmentation Approach for Deep
  Learning Based Image Classification.
\newblock \emph{CoRR} abs/2003.13048.

\bibitem[{{Xie} et~al.(2016){Xie}, {Wang}, {Wei}, {Wang}, and {Tian}}]{7780883}
{Xie}, L.; {Wang}, J.; {Wei}, Z.; {Wang}, M.; and {Tian}, Q. 2016.
\newblock DisturbLabel: Regularizing CNN on the Loss Layer.
\newblock In \emph{2016 IEEE Conference on Computer Vision and Pattern
  Recognition (CVPR)}, 4753--4762.

\bibitem[{Yang et~al.(2019)Yang, Xie, Qiao, and
  Yuille}]{DBLP:conf/aaai/YangXQY19}
Yang, C.; Xie, L.; Qiao, S.; and Yuille, A.~L. 2019.
\newblock Training Deep Neural Networks in Generations: {A} More Tolerant
  Teacher Educates Better Students.
\newblock In \emph{AAAI}, 5628--5635. {AAAI} Press.

\bibitem[{Yuan et~al.(2019)Yuan, Tay, Li, Wang, and Feng}]{yuan2019revisit}
Yuan, L.; Tay, F. E.~H.; Li, G.; Wang, T.; and Feng, J. 2019.
\newblock Revisit Knowledge Distillation: a Teacher-free Framework.
\newblock In \emph{arXiv}.

\bibitem[{Yun et~al.(2019)Yun, Han, Chun, Oh, Yoo, and Choe}]{cutmix19}
Yun, S.; Han, D.; Chun, S.; Oh, S.~J.; Yoo, Y.; and Choe, J. 2019.
\newblock CutMix: Regularization Strategy to Train Strong Classifiers With
  Localizable Features.
\newblock In \emph{ICCV}, 6022--6031. {IEEE}.

\bibitem[{Zagoruyko and Komodakis(2016)}]{DBLP:conf/bmvc/ZagoruykoK16}
Zagoruyko, S.; and Komodakis, N. 2016.
\newblock Wide Residual Networks.
\newblock In \emph{Proceedings of the British Machine Vision Conference 2016,
  {BMVC} 2016, York, UK, September 19-22, 2016}.

\bibitem[{Zhang et~al.(2018)Zhang, Ciss{\'{e}}, Dauphin, and
  Lopez{-}Paz}]{Mixup17}
Zhang, H.; Ciss{\'{e}}, M.; Dauphin, Y.~N.; and Lopez{-}Paz, D. 2018.
\newblock mixup: Beyond Empirical Risk Minimization.
\newblock In \emph{ICLR}.

\bibitem[{Zhu et~al.(2020)Zhu, Jiang, Zheng, Guo, Huang, Sun, and
  Zheng}]{DBLP:conf/aaai/ZhuJZGHSZ20}
Zhu, Z.; Jiang, X.; Zheng, F.; Guo, X.; Huang, F.; Sun, X.; and Zheng, W. 2020.
\newblock Viewpoint-Aware Loss with Angular Regularization for Person
  Re-Identification.
\newblock In \emph{AAAI}, 13114--13121.

\bibitem[{{Zoph} et~al.(2018){Zoph}, {Vasudevan}, {Shlens}, and {Le}}]{8579005}
{Zoph}, B.; {Vasudevan}, V.; {Shlens}, J.; and {Le}, Q.~V. 2018.
\newblock Learning Transferable Architectures for Scalable Image Recognition.
\newblock In \emph{2018 IEEE/CVF Conference on Computer Vision and Pattern
  Recognition}, 8697--8710.

\end{thebibliography}
\end{document}